# Shape and Spatially-Varying Reflectance Estimation From Virtual Exemplars

Zhuo Hui, *Student Member, IEEE,* and Aswin C. Sankaranarayanan, *Member, IEEE*

**Abstract**—This paper addresses the problem of estimating the shape of objects that exhibit spatially-varying reflectance. We assume that multiple images of the object are obtained under a fixed view-point and varying illumination, i.e., the setting of photometric stereo. At the core of our techniques is the assumption that the BRDF at each pixel lies in the non-negative span of a known BRDF dictionary. This assumption enables a per-pixel surface normal and BRDF estimation framework that is computationally tractable and requires no initialization in spite of the underlying problem being non-convex. Our estimation framework first solves for the surface normal at each pixel using a variant of example-based photometric stereo. We design an efficient multi-scale search strategy for estimating the surface normal and subsequently, refine this estimate using a gradient descent procedure. Given the surface normal estimate, we solve for the spatially-varying BRDF by constraining the BRDF at each pixel to be in the span of the BRDF dictionary; here, we use additional priors to further regularize the solution. A hallmark of our approach is that it does not require iterative optimization techniques nor the need for careful initialization, both of which are endemic to most state-of-the-art techniques. We showcase the performance of our technique on a wide range of simulated and real scenes where we outperform competing methods.

## 1 INTRODUCTION

Photometric stereo [1] seeks to estimate the shape of an object from images obtained from a static camera and under varying lighting. While there has been remarkable progress in photometric stereo, the vast majority of techniques are devoted to scenes that exhibit simple reflectance properties. In particular, scenes with Lambertian reflectance have received the bulk of the attention [1]–[4] due to the immense simplification that such an assumption provides. Unfortunately, real-life scenes often involve non-Lambertian materials that interact with light in complex ways; this creates a significant disconnect between theory and practice.

In this paper, we present a photometric stereo method for recovering the shape and the reflectance of opaque objects that exhibit spatially-varying reflectance. The key challenge here is that the reflectance, characterized in terms of spatially-varying bidirectional reflectance distribution function (SV-BRDF), and the shape, characterized in terms of surface normals, are inherently coupled and need to be estimated jointly. Further, the SV-BRDF is a 6D function of space and incident/outgoing angles and hence, can be very high-dimensional. In the absence of additional assumptions, estimating the SV-BRDF requires a large number of input images for robust estimation.

A common assumption for enabling computationally tractable models for SV-BRDF is that the BRDF at each pixel is a weighted combination of a *few, unknown reference BRDFs* [5]. The SV-BRDF is now represented using the reference BRDFs and their relative abundances at each pixel. This model offers a significant reduction in the dimensionality of

- *Z. Hui and A. C. Sankaranarayanan are with the Department of Electrical and Computer Engineering, Carnegie Mellon University, Pittsburgh, PA. E-mail: {zhui, saswin}@andrew.cmu.edu*
  *The authors were supported in part by the NSF grant CCF-1117939.*

the unknowns and, as a consequence, has been used in the context of photometric stereo [6], [7]. In Goldman et al. [6], the parametric isotropic Ward model [8] is used to characterize the reference BRDFs. Alldrin et al. [7] assume that the reference BRDFs are approximated by the non-parametric bivariate model [9] that approximates the 4D BRDF as a 2D signal. In both cases, the problem of shape and SV-BRDF estimation reduces to alternating minimization over the surface normals, the reference BRDFs, and abundances of the reference BRDFs at each pixel. The drawback of these approaches is that the optimization is not just computationally expensive but also has a critical dependence on the ability to find a good initial solution since the underlying problem is non-convex and riddled with local minima.

An alternate approach called example-based photometric stereo [10] introduces reference objects — typically, spherical objects — in the scene. This technique relies on the concept of *orientation consistency* [10] which suggests that two surface elements with identical normals and BRDFs will take the same appearance when placed in the same illumination. Example-based photometric stereo exploits orientation consistency as follows. Suppose that we want to estimate the surface normal at a particular pixel on the target. If the reference sphere has the same BRDF as the target, then we simply compare the intensity profile observed at each pixel on the sphere to that observed on the target pixel. The surface normal at the target pixel is recovered by finding the pixel on the sphere that best matches the intensity profile. In essence, each pixel on the reference sphere provides a candidate for the true surface normal. When the target's BRDF is spatially-varying, two reference objects — one diffuse and the one specular — can be used to recover the surface normals of the target by approximating the unknown BRDF at each pixel as a non-negative linear combination of the reference BRDFs [10]. A hallmark of example-based photometric stereo is that we do not need to calibrate the illumination. While example-based photometric stereo produces precise shape estimates without requiring the knowledge of lighting, there are multiple drawbacks associated with the method. The accuracy of recovering the surface normals is affected by the non-uniform sampling of normals of the spherical objects; specifically, we can expect to observe dense sampling of candidate normals along the viewing direction and coarse sampling near the vanishing directions. Many BRDFs are also poorly approximated as a linear combination of the two reference BRDFs. Finally, introducing reference objects is not always desirable in many practical applications.

The technique proposed in this paper relies on the core principle of example-based photometric stereo *without actually introducing reference objects into the scene*. Given a dictionary whose atoms are BRDFs associated with a wide range of materials, we can render virtual spheres, one for each atom in the dictionary, under the knowledge of the scene illumination. This provides a set of "virtual exemplars" that can be used to obtain a per-pixel estimate of the shape and reflectance of the scene with arbitrary spatially-varying BRDF. The assumption that we make is that the unknown BRDF at each pixel lies in the non-negative span of the dictionary atoms. We show that the surface normals and the BRDFs can be estimated via a sequence of tractable linear inverse problems. This obviates the need for complex iterative optimization techniques as well as careful initialization required to avoid convergence to local minima. The interplay of these ideas for both the normal and SV-BRDF estimation provides not just a tractable solution to a previous ill-posed problem but also state-of-the-art results on challenging scenes (see Figure 1).

**Contributions.** We make the following contributions.

[**Model**] We propose the use of a dictionary of BRDFs to regularize the surface normal and SV-BRDF estimation. The BRDF at each pixel of an object is assumed to lie in the non-negative span of the dictionary atoms.

[**Normal estimation**] We show that the surface normal at each pixel can be efficiently estimated using a coarse-to-fine search and further refined using a gradient descent-based algorithm.

[**SV-BRDF estimation**] Given the surface normals, we first recover the BRDF at each pixel independently by solving a linear inverse problem that enforces sparsity in the occurrence of the reference BRDFs at the pixel. To further regularize the BRDF estimation and obtain estimates with improved accuracy, we impose a low rank constraint on the SV-BRDF.

[**Validation**] We showcase the accuracy of the shape and SV-BRDF estimation technique on a wide range of simulated and real scenes and demonstrate that the proposed technique outperforms state-of-the-art.

A short version of this paper appeared in [11]. We have since improved the core ideas in two important ways. First, in [11], the precision of surface normal estimation is inherently limited by the sampling of the candidate normals. We have addressed this limitation by performing gradient descent to improve the precision of surface normal estimates while adding little to the time required for estimation. Second, [11] uses a non-parametric SV-BRDF estimation algorithm where the BRDF at each pixel is independently recovered; while per-pixel BRDF estimation is desirable, the recovered estimation is often erroneous due to limited amount of information available at each pixel. To alleviate this problem, we introduce an additional constraint that limits the number of unique reflectance functions in the scene; we achieve this by enforcing a low rank constraint to regularize the SV-BRDF. While the use of low-rank priors for SV-BRDF estimation is inspired in part by prior works [5]–[7], our optimization framework is significantly more tractable. We show that incorporating this low-rank prior leads to a SV-BRDF estimation that is more robust.

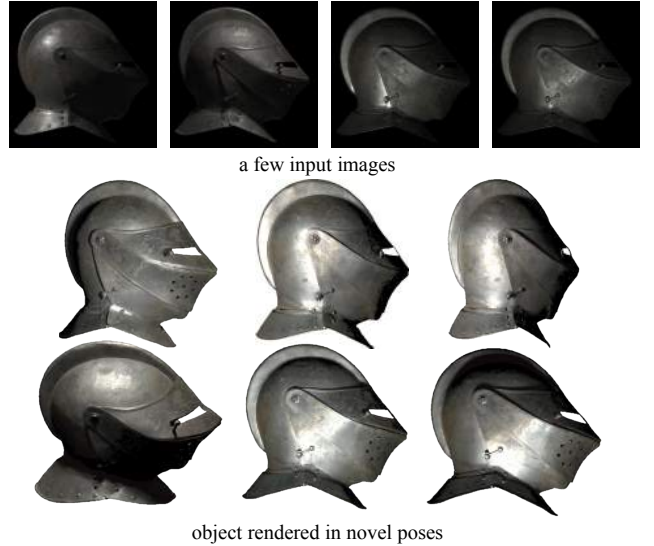

Fig. 1: **Recovery of surface normals and spatially-varying BRDF.** We propose a framework for per-pixel estimation of surface normal and BRDF in the setting of photometric stereo. Shown above are the estimated shape and rendered images of a visually-complex object. The results were obtained from 250 images.

## 2 PRIOR WORK

In this section, we review some of the key techniques for shape estimation with respect to different BRDF models.

**The diffuse + specular BRDF model.** It is well known that the collection of images of a convex Lambertian object typically lies close to a low-dimensional subspace [12], [13]. This naturally leads to techniques [14]–[16] that robustly fit a low-dimensional subspace, capturing the Lambertian component while isolating non-Lambertian components, such as specularities, as sparse outliers. From the low-dimensional space reconstructed, they implement Lambertain photometric stereo to get the shape of objects. However, these techniques have restrictive assumptions on the range of BRDFs to which they are applicable, and more importantly, miss out on powerful cues to the shape of the object that are often present in specular highlights.

**Parametric BRDF representations.** Parametric models such as the Blinn-Phong [17], Ward [8], Oren-Nayar [18], Ashikhmin-Shirley [19], Lafortune et al. [20], He et al. [21]

and Cook-Torrance model [22] are based on macro-behavior established using specific micro-facet models on the materials, and have been widely used in computer graphics. In the context of shape and SV-BRDF estimation, Goldman et al. [6] utilize the isotropic Ward model [8] to reduce the dimensionality of the inverse problem. Oxholm and Nishino [23]–[25] further extend this idea by introducing a probabilistic formulation to estimate the BRDFs and exploit visual cues from multiple views under natural lighting conditions to reconstruct the object's shape. However, parametric models are inherently limited in their ability to provide precise approximations to the true BRDFs and further, often lead to challenging and ill-conditioned inverse problems.

**Non-parametric BRDF representations.** Non-parametric models are built upon the raw measured BRDFs [26], [27] and can provide faithful rendition to the empirical observations. The BRDFs are tabulated with respect to four angles, two for the incident direction and two for the outgoing direction. The high-dimensionality of non-parametric BRDF representations is often a challenge when we need to perform BRDF estimation, even when the shape is known.

**Isotropic BRDFs.** Isotropic materials exhibit a form of symmetry, wherein the reflectance of the material is unchanged when the incident and outgoing directions are jointly rotated about the surface normal. This enables the representation of isotropic BRDFs as the function over three as opposed to four angles. In the context of photometric stereo, Alldrin and Kriegman [28] observe that, for isotropic materials, the surface normal at each point can be restricted to lie on a plane. By restricting the light source with circular motion, Chandraker et al. [29] show that the shape can be estimated from the iso-contours of depth as well as an initial starting surface normal. When the isotropic BRDF has a single dominant lobe, Shi et al. [30] resolve the planar ambiguity and show that the surface normals can be uniquely determined. For the materials with multiple lobes, Shi et al. [31] address the problem by utilizing biquadratic to characterize the low-frequency components of isotropic materials, allowing for the normal estimation via solving a least square problem from the diffuse components. Ikehata and Aizawa [32] model the isotropic BRDFs as the sum of bivariate functions and solve for the surface normals via a constrained regression problem. Higo et al. [33] utilize properties of isotropy, visibility and monotonicity to restrict the solution space of the surface normal at each pixel. This enables a framework for shape estimation without the need for radiometric calibration. Lu et al. [34]–[36] further extend the idea by exploiting the relation between surface normals and observed intensity profiles to estimate the shape of the object from multiple images without illumination calibration. Finally, a bivariate approximation for isotropic materials is used in Romeiro et al. [9], [37] to estimate the BRDF of a known shape from a single image and without knowledge of the scene illumination.

**Relationship to prior work.** There have been other methods similar to our approach that seek to remove the use of "examples" from example-based photometric stereo. In Ackermann et al. [38], [39], a partial reconstruction of the scene using multi-view stereo techniques is used as a reference (or example) to obtain dense normal estimates. In contrast, our technique focuses on the traditional problem of single-view photometric stereo. The assumption of the scene's reflectance function being composed of a few reference BRDFs is a common assumption used for photometric stereo under SV-BRDFs [5]–[7], [40], [41]. However, this leads to a multi-linear optimization in high-dimensional variables (the reference BRDFs) that is highly dependent on initial conditions. In contrast, our proposed technique avoids the need to estimate high-dimensional optimization by evoking knowledge of a dictionary of BRDFs.

## 3 PROBLEM SETUP

**Setup.** We make the following assumptions, most of which are typical to photometric stereo-based shape estimation. First, the camera is orthographic and hence, the viewing direction $\mathbf{v} \in \mathbb{R}^3$ is constant across all scene points. Second, the scene illumination is assumed to be from a distant point light source. The light sources are assumed to be of constant brightness (equivalently, that calibration is known) and their direction is known. We denote $\mathbf{l}_k \in \mathbb{R}^3$ to refer to the lighting direction in the $k$-th image $I^k$. For a light-stage, this information is typically obtained by a one-off calibration. Third, the effects of long-range illumination such as cast shadows and inter-reflections are assumed to be negligible; this is satisfied for objects with a convex shape. Finally, the radiometric response of the camera is linear.

**BRDF representation.** We follow the isotropic BRDF representation used in [42] in which a three-angle coordinate system based on half angles is used. Specifically, the BRDF is expressed as a function $\rho(\theta_h, \theta_d, \phi_d)$ with $\theta_h, \theta_d \in [0, \pi/2)$ and $\phi_d \in [0, 2\pi)$. However, by Helmholtz's reciprocity, the BRDF exhibits the following symmetry: $\rho(\theta_h, \theta_d, \phi_d) = \rho(\theta_h, \theta_d, \phi_d + \pi)$, and hence it is sufficient to express $\phi_d \in [0, \pi)$. Following [26], we use a 1° sampling of each angle. As a consequence, a BRDF is represented as a point in a $T = 90 \times 90 \times 180 = 1,458,000$-dimensional space. When we deal with color images, we have a BRDF for each color channel and hence, the dimensionality of the BRDF goes up proportionally.

Consider a scene element with BRDF $\rho \in \mathbb{R}^T$, surface normal $\mathbf{n}$, illuminated by a point light source from a direction $\mathbf{l}$ and viewed from a direction $\mathbf{v}$. For this configuration of normal, incident light and viewing direction, the BRDF value is simply a linear functional of the vector $\rho$:

$$\mathbf{s}_{\{\mathbf{l},\mathbf{v};\mathbf{n}\}}^\top \rho,$$

where $\mathbf{s}_{\{\mathbf{l},\mathbf{v};\mathbf{n}\}}$ is a vector that encodes the geometry of the configuration. In essence, the vector samples the appropriate entry from $\rho$, allowing for the appropriate interpolation if the required value is off the sampling-grid.

**Problem formulation.** Our goal is to recover the surface normals and the SV-BRDF in the context of photometric stereo; i.e., multiple images of an object $\{I^1, \ldots, I^Q\}$ obtained from a static camera under varying lighting. The intensity value $I_\mathbf{p}^i$ observed at pixel $\mathbf{p} = (x, y)$ with lighting $\mathbf{l}_i$ can be written as

$$I_\mathbf{p}^i = (\mathbf{s}_{\{\mathbf{l}_i,\mathbf{v};\mathbf{n}_\mathbf{p}\}}^\top \rho_\mathbf{p}) \cdot \max\{0, \mathbf{n}_\mathbf{p}^\top \mathbf{l}_i\}, \qquad (1)$$

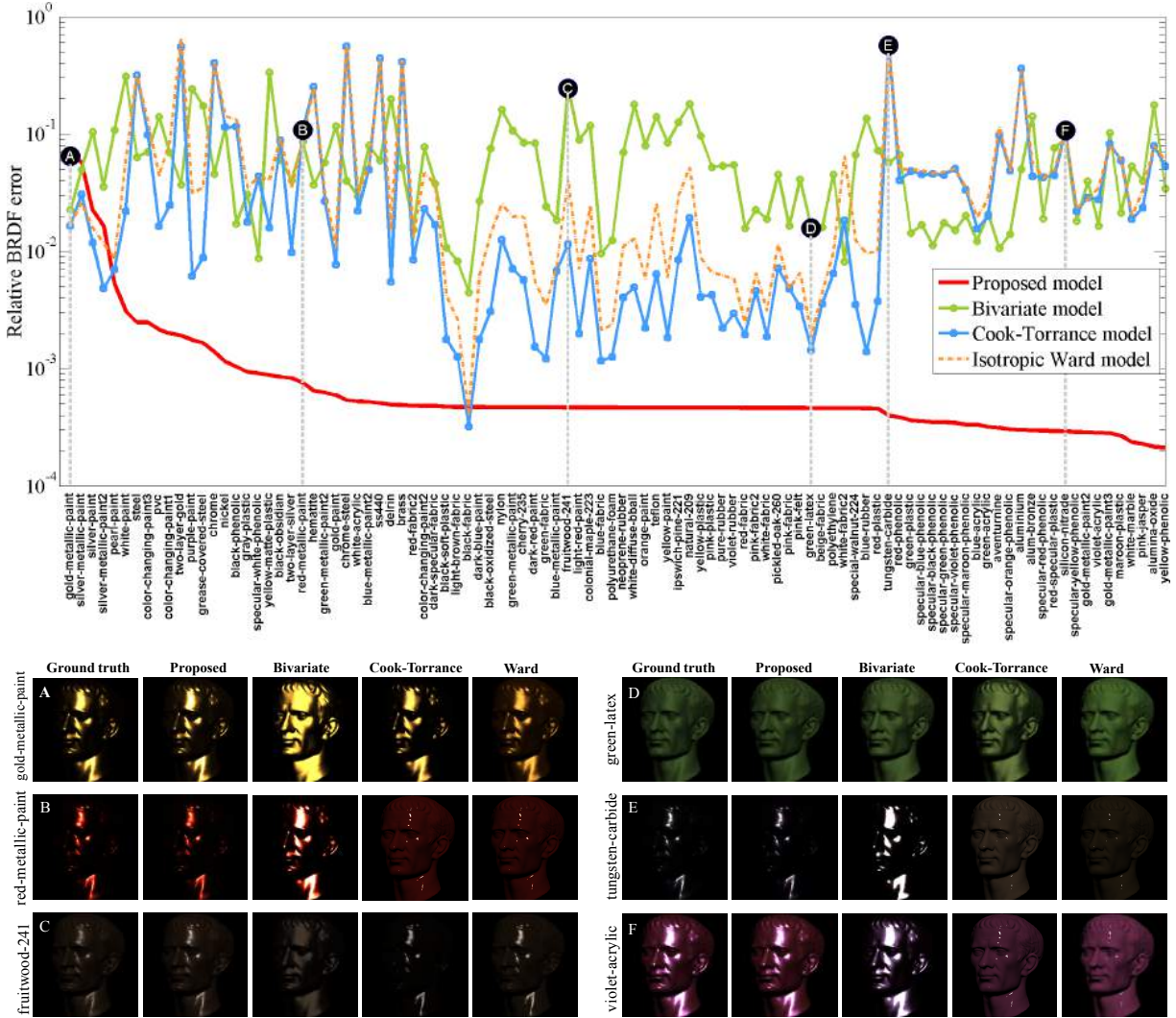

Fig. 2: **Accuracy of BRDF models on the MERL database [26].** For the 100 materials in the database, we plot the approximation accuracy in relative RMS error [27] (also see (8)) for the proposed, bivariate [9], Cook-Torrance [22], and the isotropic Ward [8] models. For the proposed model, we use a leave-one-out scheme, wherein for each BRDF the remaining 99 BRDFs in the database are used to form the dictionary. The proposed model outperforms competing models both quantitatively (top) as well as in visual perception (bottom).

where $\rho_\mathbf{p}$ is the BRDF and $\mathbf{n_p}$ is the surface normal at pixel $\mathbf{p}$, respectively, and $\max\{0, \mathbf{n_p}^\top \mathbf{l}_i\}$ accounts for shading.

Given multiple intensity values at pixel $\mathbf{p}$, one for each lighting direction $\{\mathbf{l}_1, \ldots, \mathbf{l}_Q\}$, we can write

$$\begin{aligned} \mathbf{I_p} &= \begin{pmatrix} I_\mathbf{p}^1 \\ \vdots \\ I_\mathbf{p}^Q \end{pmatrix} = \begin{bmatrix} \max\{0, \mathbf{n_p}^\top \mathbf{l}_1\} \cdot \mathbf{s}^\top_{\{\mathbf{l}_1, \mathbf{v}; \mathbf{n_p}\}} \\ \vdots \\ \max\{0, \mathbf{n_p}^\top \mathbf{l}_Q\} \cdot \mathbf{s}^\top_{\{\mathbf{l}_Q, \mathbf{v}; \mathbf{n_p}\}} \end{bmatrix} \rho_\mathbf{p}, \\ &= A(\mathbf{n_p}) \rho_\mathbf{p}. \end{aligned} \quad (2)$$

Given the intensities, $\mathbf{I_p}$, observed at a pixel $\mathbf{p}$ and knowledge of lighting directions $\{\mathbf{l}_1, \ldots, \mathbf{l}_Q\}$, we seek to estimate the surface normal $\mathbf{n_p}$ and the BRDF $\rho_\mathbf{p}$ at the pixel. This problem is intractable without additional assumptions that constrain the BRDF to a lower-dimensional space.

**Model for BRDF.** The key assumption that we make is that the BRDF at a pixel $\mathbf{p}$ lies on the non-negative span of the atoms of a BRDF dictionary. Specifically, given dictionary $D = [\rho^1, \rho^2, \cdots, \rho^M]$, we assume that the BRDF at pixel $\mathbf{p}$ can be written as

$$\rho_\mathbf{p} = D \mathbf{c_p}, \quad \mathbf{c_p} \geq 0,$$

where $\mathbf{c_p} \in \mathbb{R}^M$ are the abundances of the dictionary atoms. In essence, we have constrained the BRDF to lie in an $M$-dimensional cone.[1] This provides immense reduction in the dimensionality of the unknowns at the expense of introducing a model misfit error. Indeed the success of this model relies on having a dictionary that is sufficiently rich to cover a wide range of interesting materials. Figure 2 shows the accuracy of various BRDF models on the MERL BRDF database [26].

---

1. A more appropriate model for the BRDF is that $(D\mathbf{c}) \geq 0$. However, this leads to significantly higher-dimensional constraints. We instead use a sufficient condition to achieve this, $\mathbf{c} \geq 0$.

In addition to the dictionary model for the BRDF, we also consider two additional priors.

- *Sparsity.* In the context of per-pixel BRDF estimation, we assume that $\mathbf{c_p}$ is sparse, suggesting that BRDF at the pixel $\mathbf{p}$ is the linear combination of a *few* dictionary atoms. The sparsity constraint avoids over-fitting to the intensity measurements $\mathbf{I_p}$ as well as provides a regularization for under-determined problems.
- *Low rank.* In the context of estimating the SV-BRDF for all pixels jointly, we assume that coefficient matrix $\mathbf{C} = [\mathbf{c_{p_1}}, \mathbf{c_{p_2}}, \ldots \mathbf{c_{p_N}}]$, that denotes the collection of the abundances for all the $N$ pixels in the scene, is low rank. The low-rank prior on $\mathbf{C}$ implies that BRDFs at all pixels can be expressed as a linear combination of small number of unique reflectance functions. This prior is at the heart of many approaches for photometric stereo under SV-BRDF [5]–[7], [41]. The low-rank prior also enables us to efficiently pool together information from multiple pixels, thereby providing significant improvements over the per-pixel estimates, without exploiting any explicit spatial smoothness priors.

**Solution outline.** We formulate the per-pixel surface normal and BRDF estimation using the following optimization problem.

$$\{\widehat{\mathbf{n}}_\mathbf{p}, \widehat{\mathbf{c}}_\mathbf{p}\} = \arg\min_{\mathbf{n},\mathbf{c}} \|\mathbf{I_p} - A(\mathbf{n})D\mathbf{c}\|_2^2 + \lambda\|\mathbf{c}\|_1 \quad (3)$$
$$\text{s.t} \quad \mathbf{c} \geq 0, \|\mathbf{n}\|_2 = 1.$$

The $\ell_1$-penalty serves to enforce sparse solutions, with $\lambda \geq 0$ determining the level of sparsity in the solution. The optimization problem in (3) is non-convex due to unit-norm constraint on the surface normal $\mathbf{n}$ as well as the term $A(\mathbf{n})D\mathbf{c}$. Our solution methodology consists of two steps:

1) *Surface normal estimation.* We perform an efficient multi-scale search together with the gradient descent based refinement scheme which provides us with a precise estimate of the surface normal at pixel $\mathbf{p}$ (see Section 4);
2) *BRDF estimation.* We first solve (3) only over $\mathbf{c}$ with the normal fixed to obtain the BRDF at $\mathbf{p}$. We next incorporate a low rank constraint on the SV-BRDF to further regularize the BRDF estimates (see Section 5).

## 4 SURFACE NORMAL ESTIMATION

In this section, we describe an efficient per-pixel surface normal estimation algorithm.

### 4.1 Virtual exemplar-based normal estimation

The first step of our surface normal estimation can be viewed as an extension of the method proposed in [10], where two spheres — one diffuse and one specular — are introduced in a scene along with the target object. Recall that, the scene is observed under $Q$ different illuminations. Hence, at a pixel $\mathbf{p}$ on the target, we can construct the intensity profile $\mathbf{I_p} \in \mathbb{R}^Q$ that enumerates the $Q$ intensity values observed at $\mathbf{p}$. To obtain the surface normals at the pixel $\mathbf{p}$, we compare the intensity profile, $\mathbf{I_p}$, to those on the reference spheres. The reference spheres provide a sampling of the space of the normals and hence, we can simply treat them as a collection of candidate normals $\mathcal{N}$.

By orientation consistency, the surface normal estimation now reduces to finding the candidate normal that can best explain the intensity profile $\mathbf{I_p}$. Given a candidate normal $\widetilde{\mathbf{n}}$, we have two intensity profiles, $\mathbf{I}_D(\widetilde{\mathbf{n}})$ and $\mathbf{I}_S(\widetilde{\mathbf{n}})$, one each for the diffuse and specular sphere, respectively. The estimate of the surface normal at pixel $\mathbf{p}$ is given as

$$\widehat{\mathbf{n}}_\mathbf{p} = \arg\min_{\widetilde{\mathbf{n}} \in \mathcal{N}} \min_{a_1, a_2 \geq 0} \|\mathbf{I_p} - a_1 \mathbf{I}_D(\widetilde{\mathbf{n}}) - a_2 \mathbf{I}_S(\widetilde{\mathbf{n}})\|.$$

In [10], this is solved by scanning over all the pixels/candidate normals on the reference spheres.

**Rendering virtual spheres.** We rely on the same approach as [10] with the key difference that we virtually render the reference spheres. The virtual spheres are rendered as follows. Given the lighting directions $\{\mathbf{l}_1, \ldots, \mathbf{l}_Q\}$ and the BRDF dictionary $D = [\rho^1, \ldots, \rho^M]$, for each candidate normal $\widetilde{\mathbf{n}} \in \mathcal{N}$, we render a matrix $B(\widetilde{\mathbf{n}}) = [b_{ij}(\widetilde{\mathbf{n}})] \in \mathbb{R}^{Q \times M}$ such that $b_{ij}(\widetilde{\mathbf{n}})$ is the intensity observed at a surface with normal $\widetilde{\mathbf{n}}$ and BRDF $\rho^j$, under lighting $\mathbf{l}_i$.

$$b_{ij}(\widetilde{\mathbf{n}}) = \max\{0, \widetilde{\mathbf{n}}^\top \mathbf{l}_i\} \cdot \mathbf{s}_{\{\mathbf{l}_i, \mathbf{v}; \widetilde{\mathbf{n}}\}}^\top \rho^j,$$

We render one such matrix $B(\cdot)$ for each candidate normal in $\mathcal{N}$. Given these virtually rendered spheres, we can solve (3) by searching over all candidate normals.

**Brute-force search.** For computationally efficiency, we drop the sparsity-promoting term in (3). We empirically observed that this makes little difference in the estimated surface normals. Now, given the intensity profile $\mathbf{I_p}$ at pixel $\mathbf{p}$ and noting that $B(\widetilde{\mathbf{n}}) = A(\widetilde{\mathbf{n}})D$, solving (3) reduces to:

$$\widehat{\mathbf{n}}_\mathbf{p} = \arg\min_{\widetilde{\mathbf{n}} \in \mathcal{N}} \min_{\mathbf{c} \geq 0} \|\mathbf{I_p} - B(\widetilde{\mathbf{n}})\mathbf{c}\|_2^2. \quad (4)$$

The unit-norm constraint on the surface normals is absorbed into the candidate normals being unit-norm. The optimization problem in (4) requires solving a set of non-negative least squares (NNLS) sub-problems, one for each element of $\mathcal{N}$. For the results in the paper, we used the `lsnonneg` function in MATLAB to solve the NNLS sub-problems.

The accuracy and the computational cost in solving (4) depends solely on the cardinality of the candidate set $\mathcal{N}$, $|\mathcal{N}|$. We obtain $\mathcal{N}$ by uniform or equi-angular sampling on the sphere [43]. Note that the smaller the angular spacing of $\mathcal{N}$, the larger is its cardinality. For example, a $5°$ equi-angular sampling over the hemisphere requires approximately 250 candidates, while a $0.5°$ requires 20,000 candidates. Given that the time-complexity of the brute-force search is linear in $|\mathcal{N}|$, the computational costs for obtaining very precise normal estimates can be overwhelming (see Table 1). To alleviate this, we outline a coarse-to-fine search strategy that is remarkably faster than the brute-force approach with little loss in accuracy.

### 4.2 Coarse-to-fine search

Figure 3 shows the value of

$$E(\widetilde{\mathbf{n}}) = \min_{\mathbf{c} \geq 0} \|\mathbf{I_p} - B(\widetilde{\mathbf{n}})\mathbf{c}\|$$

as a function of the candidate normal $\widetilde{\mathbf{n}}$ for a few examples. We observe that there is a gradual increase in error value as we moved away from the global minima of $E(\cdot)$. We exploit this to design a coarse-to-fine search strategy where we first evaluate the candidate normals at a coarse sampling and subsequently search in the vicinity of this solution at a finer sampling.

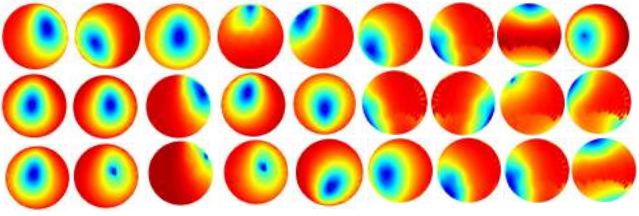

Fig. 3: **The error as a function of candidate normals for a few test examples**. We can observe that the global minima is compact and the error increases largely monotonically in its vicinity. This motivates our coarse-to-fine search strategy.

|  |  | $\theta_1$ | $\theta_2$ | $\theta_3$ | $\theta_4$ | $\theta_5$ |
|---|---|---|---|---|---|---|
| **Brute force** | time | 0.18s | 0.77s | 4.25s | 27.3s | 74.1s |
|  | ang. error | 7.07° | 3.99° | 1.56° | 0.60° | 0.42° |
|  | max samples | 76 | 327 | 1828 | 11829 | 31830 |
| **Coarse to fine** | time | 0.18s | 0.19s | 0.23s | 0.34s | 0.41s |
|  | ang. error | 7.07° | 4.99° | 2.56° | 1.23° | 0.82° |
|  | max samples | 76 | 81 | 89 | 105 | 112 |

TABLE 1: Comparison of brute-force and coarse-to-fine normal estimation for different angular samplings of the candidate normals: $\theta_1 = 10°, \theta_2 = 5°, \theta_3 = 3°, \theta_4 = 1°$, and $\theta_5 = 0.5°$. For each method, we report the time taken, the angular error, and the maximum number of candidates evaluated. Shown are averages over 100 random trials.

Specifically, let $\mathcal{N}_\theta$ be the set of equi-angular sampling on the unit-sphere where the angular spacing is $\theta$ degrees. Given a candidate normal $\widetilde{\mathbf{n}}$, we define

$$C_\theta(\widetilde{\mathbf{n}}) = \{\mathbf{n} \mid \langle \mathbf{n}, \widetilde{\mathbf{n}} \rangle \geq \cos\theta, \ \|\mathbf{n}\|_2 = 1\}$$

as the set of unit-norm vectors within $\theta$-degrees from $\widetilde{\mathbf{n}}$.

In the first iteration, we initialize the candidate normal set $\mathcal{N}^{(1)} = \mathcal{N}_{\theta_1}$. Now, at the $j$-th iteration, we solve (4) over a candidate set $\mathcal{N}^{(j)}$. Suppose that $\widehat{\mathbf{n}}^{(j)}$ is the candidate normal where the minimum occurs at the $j$-th iteration. The candidate set for the $(j+1)$-th iteration is constructed as

$$j \geq 1, \quad \mathcal{N}^{(j+1)} = C_{\theta_j}(\widehat{\mathbf{n}}^{(j)}) \cap \mathcal{N}_{\theta_{j+1}},$$

with $\theta_{j+1} < \theta_j$. That is, the candidate set is simply the set of all candidates at a finer angular sampling that are no greater than the current angular sampling from the current estimate. This is repeated till we reach the finest resolution at which we have candidate normals. For the results in this paper, we use the following values: $\theta_1 = 10°, \theta_2 = 5°, \theta_3 = 3°, \theta_4 = 1°$, and $\theta_5 = 0.5°$. For efficient implementation, we pre-render $B(\widetilde{\mathbf{n}})$ for $\widetilde{\mathbf{n}} \in \mathcal{N}_{\theta_1} \cup \cdots \cup \mathcal{N}_{\theta_5}$.

The computational gains obtained via this coarse-to-fine search strategy are immense. Table 1 shows the run-time and precision of both brute force and coarse-to-fine normal estimation strategy for different levels of angular sampling in the generation of the candidate normal set. As expected the run-time of the brute force algorithm is linear in the number of candidates. In contrast, the coarse-to-fine strategy requires a tiny fraction of this time while nearly achieving the same precision as the brute force strategy.

A drawback of both the brute-force as well as the coarse-to-fine approaches is that the estimated normals are restricted by the candidate normal set and hence, the accuracy of the estimates, on an average, cannot be better than the half the angular spacing of the candidate set at the finest level. To address this, we propose a local descent-based scheme that circumvents the limitations of using just candidate normals.

### 4.3 Gradient descent-based normal estimation

Our gradient descent-based scheme to estimate the surface normals relies on two observations: first, we can use the estimate obtained from the coarse-to-fine strategy as an accurate initial guess; and second, we can linearize the cost function in (4) in the vicinity of our initial guess and devise a gradient descent algorithm.

Specifically, let $f(\mathbf{n_p}, \mathbf{c_p})$ be the value of the data term in (4), i.e.

$$f(\mathbf{n_p}, \mathbf{c_p}) = \|\mathbf{I_p} - B(\mathbf{n_p})\mathbf{c_p}\|_2^2.$$

Now, at $\widehat{\mathbf{n}}_\mathbf{p}, \widehat{\mathbf{c}}_\mathbf{p}$ obtained by using coarse-to-fine search, we can linearize $f(\mathbf{n_p}, \mathbf{c_p})$ as

$$f(\widehat{\mathbf{n}}_\mathbf{p}+\triangle\mathbf{n_p}, \widehat{\mathbf{c}}_\mathbf{p}+\triangle\mathbf{c_p}) = \|\mathbf{I_p} - B(\widehat{\mathbf{n}}_\mathbf{p} + \triangle\mathbf{n_p})(\widehat{\mathbf{c}}_\mathbf{p}+\triangle\mathbf{c_p})\|_2^2.$$

Given $B(\widehat{\mathbf{n}}_\mathbf{p})$ is locally smooth[2], it can be linearized at $\widehat{\mathbf{n}}_\mathbf{p}$ as

$$B(\widehat{\mathbf{n}}_\mathbf{p} + \triangle\mathbf{n_p}) = B(\widehat{\mathbf{n}}_\mathbf{p}) + \nabla_\mathbf{n} B(\widehat{\mathbf{n}}_\mathbf{p})\triangle\mathbf{n_p}.$$

To account for the unit norm constraint on $\widehat{\mathbf{n}}_\mathbf{p} + \triangle\mathbf{n_p}$, we utilize the elevation angle, which is denoted as $\theta$, and the azimuth angle, which is denoted as $\phi$, to represent surface normals. That is, we restrict the update of surface normals into a two dimensional space by absorbing the unit norm constraint. In particular, we can write $B(\widehat{\mathbf{n}}_\mathbf{p} + \triangle\mathbf{n_p})$ as

$$B(\widehat{\mathbf{n}}_\mathbf{p} + \triangle\mathbf{n_p}) = B(\widehat{\mathbf{n}}_\mathbf{p}) + \nabla_\phi B(\widehat{\mathbf{n}}_\mathbf{p})\triangle\phi_\mathbf{p} + \nabla_\theta B(\widehat{\mathbf{n}}_\mathbf{p})\triangle\theta_\mathbf{p},$$

where $\triangle\phi_\mathbf{p}$ and $\triangle\theta_\mathbf{p}$ denote local gradients for the elevation and azimuth angles of $\widehat{\mathbf{n}}_\mathbf{p}$, respectively. In essence, we have now reformulated the problem in (4) into a form involving the local gradients in surface normals and abundances. This enables us to refine the normal estimates without any restrictions imposed by the sampling of the candidate set.

Now, an estimate of local gradients at a pixel $\mathbf{p}$ can be obtained by solving

$$\{\triangle\widehat{\theta}_\mathbf{p}, \triangle\widehat{\phi}_\mathbf{p}, \triangle\widehat{\mathbf{c}}_\mathbf{p}\} = \underset{\triangle\theta, \triangle\phi, \triangle c}{\arg\min}$$
$$\|\mathbf{I_p} - (B(\widehat{\mathbf{n}}_\mathbf{p}) + \nabla_\phi B(\widehat{\mathbf{n}}_\mathbf{p})\triangle\phi + \nabla_\theta B(\widehat{\mathbf{n}}_\mathbf{p})\triangle\theta)(\widehat{\mathbf{c}}_\mathbf{p} + \triangle\mathbf{c})\|_2^2$$
$$\text{s.t} \quad \widehat{\mathbf{c}}_\mathbf{p} + \triangle\mathbf{c} \geq 0. \tag{5}$$

We drop the second-order terms $\triangle\theta\triangle\mathbf{c}$ and $\triangle\phi\triangle\mathbf{c}$ in (5), which contributes little energy to the cost function, and we can solve the resulting convex optimization problem in (5) over $\triangle\theta$, $\triangle\phi$ and $\triangle\mathbf{c}$ using alternating minimization.

---

2. Though $B(\mathbf{n})$ involves the shading term, for the small angular spacing, like $0.5°$ in this paper, the smoothness can also hold for most of candidate normals.

**Estimating $\nabla_\phi B(\widehat{\mathbf{n}}_\mathbf{p})$ and $\nabla_\theta B(\widehat{\mathbf{n}}_\mathbf{p})$.** Let $\mathcal{N}^{(J)}$ be the candidate normals set at the finest sampling level $J$. To estimate the gradients at the current estimate $\widehat{\mathbf{n}}_\mathbf{p}$, we construct a set $\mathcal{S} \subset \mathcal{N}^{(J)}$ of all normals in $\mathcal{N}^{(J)}$ that lie in a small neighborhood (smaller than 2 degrees in angular difference) of $\widehat{\mathbf{n}}_\mathbf{p}$. For each normal $\widetilde{\mathbf{n}} \in \mathcal{S}$ we can write

$$B(\widetilde{\mathbf{n}}) - B(\widehat{\mathbf{n}}_\mathbf{p}) = (\widetilde{\phi} - \widehat{\phi}_\mathbf{p})\nabla_\phi B(\widehat{\mathbf{n}}_\mathbf{p}) + (\widetilde{\theta} - \widehat{\theta}_\mathbf{p})\nabla_\theta B(\widehat{\mathbf{n}}_\mathbf{p}),$$

where $(\widetilde{\theta}, \widetilde{\phi})$ is the Euler angle representation of $\widetilde{\mathbf{n}}$. We can set up an over-determined set of equations by stacking together the constraints arising from normals in the set $\mathcal{S}$ to estimate the gradients, $\nabla_\phi B(\widehat{\mathbf{n}}_\mathbf{p})$ and $\nabla_\theta B(\widehat{\mathbf{n}}_\mathbf{p})$. We recover the gradients by taking the pseudo-inverse of this overdetermined linear system.

Given the estimated $\nabla_\phi B(\widehat{\mathbf{n}}_\mathbf{p})$ and $\nabla_\theta B(\widehat{\mathbf{n}}_\mathbf{p})$, we perform the following steps until convergence.

**Updating $\triangle\widehat{\phi}_\mathbf{p}$ and $\triangle\widehat{\theta}_\mathbf{p}$.** Both $\triangle\widehat{\phi}$ and $\triangle\widehat{\theta}$ are estimated by solving a least square problem.

**Update $\triangle\widehat{\mathbf{c}}_\mathbf{p}$.** Due to the non-negative constraint on $\widehat{\mathbf{c}}_\mathbf{p} + \triangle\mathbf{c}$, we first solve the least square problem over $\triangle\mathbf{c}$ and then project the solution to the space specified by the constraint.

The estimate of surface normals can be obtained by

$$\widehat{\theta}_\mathbf{p} \leftarrow \widehat{\theta}_\mathbf{p} + \triangle\widehat{\theta}_\mathbf{p},$$
$$\widehat{\phi}_\mathbf{p} \leftarrow \widehat{\phi}_\mathbf{p} + \triangle\widehat{\phi}_\mathbf{p},$$
$$\widehat{\mathbf{n}}_\mathbf{p} \leftarrow [\cos(\widehat{\phi}_\mathbf{p})\sin(\widehat{\theta}_\mathbf{p}), \sin(\widehat{\phi}_\mathbf{p})\sin(\widehat{\theta}_\mathbf{p}), \cos(\widehat{\theta}_\mathbf{p})]^\top.$$

**Observations.** The gradient descent procedure described above can be solved efficiently. For a single surface normal, optimization to converge takes between 0.8 and 0.9 seconds in MATLAB on a desktop with Intel Xeon 3.6G CPU. In Table 2, we tabulate the improvements provided by the gradient descent procedure when initialized with the solutions of the brute force as well as the coarse-to-fine strategies. We observe that both algorithms benefit immensely from utilizing the gradient descent search. Further, the average error can be made smaller than the sampling resolution of candidate normals on the unit sphere. However, that the average angular error does not reduce to zero potentially due to model misfits introduced by shading term in matrix $B$ as well as the empirical estimation of the gradients.

## 5 REFLECTANCE ESTIMATION

### 5.1 Per-pixel BRDF estimate

Given the surface normal estimate $\widehat{\mathbf{n}}_\mathbf{p}$, we obtain an estimate of the BRDF at each pixel, individually, by solving

$$\widehat{\mathbf{c}}_\mathbf{p} = \arg\min_{\mathbf{c}\geq 0} \|\mathbf{I}_\mathbf{p} - B(\widehat{\mathbf{n}}_\mathbf{p})\mathbf{c}\|_2^2 + \lambda\|\mathbf{c}\|_1. \quad (6)$$

The use of the $\ell_1$-regularizer promotes sparse solutions and primarily helps in avoiding over-fitting to the observed intensities. The optimization problem in (6) is convex and we used CVX [44], a general purpose convex solver, to obtain solutions. The estimate of the BRDF at pixel $\mathbf{p}$ is given as $\widehat{\rho}_\mathbf{p} = D\widehat{\mathbf{c}}_\mathbf{p}$. The value of $\lambda$ was manually tuned for best performance on synthetic data. For color-imagery, we solve for the coefficients associated with each color channel separately.

|  |  | $\theta_1$ | $\theta_2$ | $\theta_3$ | $\theta_4$ | $\theta_5$ |
|---|---|---|---|---|---|---|
| **Brute force** | original | 7.07° | 3.99° | 1.56° | 0.60° | 0.42° |
|  | refine | 4.10° | 2.43° | 0.97° | 0.43° | 0.21° |
| **Coarse to fine** | original | 7.07° | 4.99° | 2.56° | 1.23° | 0.82° |
|  | refine | 4.10° | 2.86° | 1.57° | 0.75° | 0.38° |

TABLE 2: Gradient descent local search starting from both the brute-force and coarse-to-fine normal estimation for different angular samplings in the candidate normals: $\theta_1 = 10°, \theta_2 = 5°, \theta_3 = 3°, \theta_4 = 1°$, and $\theta_5 = 0.5°$. Shown are aggregate statistics over 100 randomly generated trials.

The advantage of the per-pixel BRDF estimation framework is the ability to handle arbitrarily complex spatial variations in the BRDF. However, a drawback of per-pixel BRDF estimation is the relative lack of information available at a pixel. When we know a priori that multiple pixels share the same BRDF, then we can solve (6) simply by concatenating their corresponding intensity profiles and their respective $B(\cdot)$ matrices. As is to be expected, pooling intensities observed at multiple pixels significantly improves the quality of the estimates. Yet, while spatial averaging or spatial smoothness priors improve the quality of the estimate, inherently they require the object to exhibit smooth spatial-variations in its BRDF. To address this, we pool together information across multiple pixels by utilizing the low rank prior.

### 5.2 Incorporating low rank priors

Given the matrix $\mathbf{C} = [\mathbf{c}_{\mathbf{p}_1}, \mathbf{c}_{\mathbf{p}_2}, \ldots, \mathbf{c}_{\mathbf{p}_N}]$ for the estimated abundances for all $N$ pixels in the scene, we constrain $\mathbf{C}$ to be low rank. The low rank prior, inspired by prior work [5]–[7], [45], suggests that the SV-BRDF of the scene under consideration can be generated from a small number of unique reference reflectance functions such that linear combinations of these reference BRDFs produces the BRDF at any pixel. The low rank prior also allows us to restrict the solution space for all the pixels globally without enforcing any spatial smoothness or clustering of pixels.

**Solution outline.** We can now formulate a global optimization problem that incorporates the low rank prior as follows: The estimate of the abundances of the BRDF at pixel $\mathbf{p}$ is given as

$$\widehat{\mathbf{C}} = \arg\min_{\mathbf{C}} \beta\|\mathbf{C}\|_* + \sum_\mathbf{p}\|\mathbf{I}_\mathbf{p} - B(\widehat{\mathbf{n}}_\mathbf{p})\mathbf{c}_\mathbf{p}\|_2^2 + \lambda\|\mathbf{c}_\mathbf{p}\|_1.$$
$$\text{s.t.} \quad \mathbf{C} = [\mathbf{c}_{\mathbf{p}_1}, \ldots, \mathbf{c}_{\mathbf{p}_N}], \quad \forall \mathbf{p}, \mathbf{c}_\mathbf{p} \geq 0. \quad (7)$$

where $\|\mathbf{C}\|_*$, the nuclear norm of the matrix $\mathbf{C}$, promotes low-rank solutions [46]–[48]. Note that we do not control the rank of the solution directly, but instead do so by using the penalty parameter $\beta$. This is achieved as follows: for large values of $\beta$, the nuclear norm penalty is strongly enforced and hence, we can expect the solution to be of low-rank. Similarly, small values of $\beta$ lead to solutions with larger rank. We exploit this observation as a precept in sequentially selecting $\beta$ till we find a solution of desired rank.

The objective in (7) consists of a smooth data fidelity term and two non-differentiable regularization terms, the $\ell_1$-term that promotes sparse solutions and the nuclear norm that promotes low-rank solutions. We solve this by using prox-linear or forward-backward operator splitting [49]. This results in the following algorithm.

Given the estimate in the $(k)$-th iteration, $\widehat{\mathbf{C}}^{(k)} = [\widehat{\mathbf{c}}_{\mathbf{p}_1}^{(k)}, \ldots, \widehat{\mathbf{c}}_{\mathbf{p}_N}^{(k)}]$, we perform three operations to obtain the estimate at the $(k+1)$-th iteration.

- **Gradient descent.** We perform the "forward operation" which comprises of a gradient descent on the smooth data fidelity term. Given the separable nature of the data fidelity term, we can apply this on each pixel separately.
$$\widehat{\mathbf{a}}_{\mathbf{p}}^{(k+1)} = \widehat{\mathbf{c}}_{\mathbf{p}}^{(k)} + 2tB(\widehat{\mathbf{n}}_{\mathbf{p}})^\top(\mathbf{I}_{\mathbf{p}} - B(\widehat{\mathbf{n}}_{\mathbf{p}})\widehat{\mathbf{c}}_{\mathbf{p}}^{(k)}),$$
where $t$ denotes the update step in gradient descent.

- **Soft thresholding.** We next perform the first "backward operation" corresponding to the $\ell_1$-norm and the non-negativity constraint. The associated proximal operator results in soft thresholding at each pixel, followed by a thresholding at zero to enforce non-negativity
$$\widehat{\mathbf{b}}_{\mathbf{p}}^{(k+1)} = \max\left(\mathcal{S}_{\lambda t}\left(\widehat{\mathbf{a}}_{\mathbf{p}}^{(k+1)}\right), \mathbf{0}\right),$$
where $\mathcal{S}_\tau(\cdot)$ denotes the soft-thresholding operator defined as $\mathcal{S}_\tau(\mathbf{x}) = \mathrm{sgn}(\mathbf{x})\max(|\mathbf{x}| - \tau, 0)$, and $|\cdot|$ is the absolute value.

- **Singular value thresholding.** Finally, we perform the second "backward operation" corresponding to the nuclear norm. The associated proximal operator results in a singular value thresholding step. Let $\widehat{\mathbf{B}}^{(k+1)} = [\widehat{\mathbf{b}}_{\mathbf{p}_1}^{(k+1)}, \ldots, \widehat{\mathbf{b}}_{\mathbf{p}_N}^{(k+1)}]$. Now, we can obtain $\widehat{\mathbf{C}}^{(k+1)}$ as
$$\widehat{\mathbf{C}}^{(k+1)} = U[\mathcal{S}_\beta(\sigma)]V^\top,$$
where $\widehat{\mathbf{B}}^{(k+1)} = U\mathrm{diag}(\sigma)V^\top$.

We perform the update for the matrix $\mathbf{C}$ until the convergence can be reached. In the next section, we carefully characterize the performance of our proposition using synthetic and real examples.

## 6 RESULTS

We characterize the performance of our technique using both synthetic and real datasets. We also direct the reader to the supplemental material for a complementary set of results that we could not accommodate due to space constraints. Finally, our code base is available for download from https://github.com/huizhuo1987/ICCP_DL_PS.

### 6.1 Synthetic experiments

We use the BRDFs in the MERL database [26] in a leave-one-out scheme for testing the accuracy of our proposed algorithms. Specifically, when we simulate a test object using a particular material, the dictionary is comprised of BRDFs of the remaining $M = 99$ materials from the database. We used the configuration in the light-stage described in [50] for our collection of lighting directions.

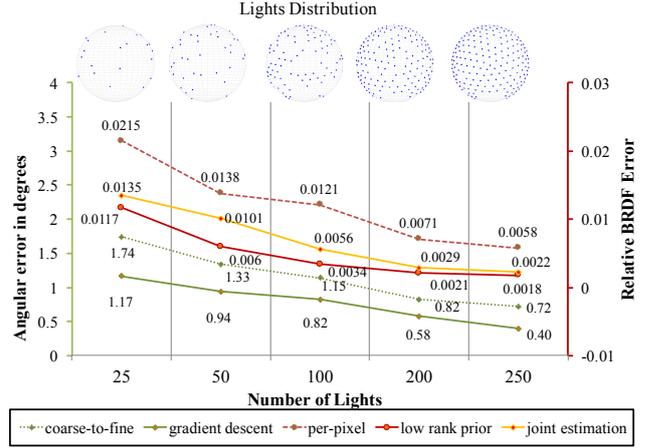

Fig. 4: **Normal and BRDF estimation with varying number of images.** We estimate the angular errors for the coarse-to-fine (in dot green) and the gradient descent method (solid green line). We also estimate the relative BRDF errors for both per-pixel (in dot red) and rank-1 prior (solid red line) under perfect knowledge of the surface normals. Finally, we test the entire estimation pipeline by measuring the accuracy of BRDFs using the surface normals from the gradient descent scheme (orange solid line). The plots were obtained by averaging across all 100 BRDFs in the MERL database and 20,000 randomly-generated normals per material.

#### 6.1.1 Performance of normal estimation

We characterize the performance of normals estimation by testing on the synthetic data with varying number of lighting directions, varying BRDFs as well as varying dictionary size and type.

**Varying number of input images.** Figure 4 characterizes the errors in surface normal for varying number of input images or equivalently, lighting directions. We report the average angular error for both the coarse-to-fine search strategy and gradient descent method. In each case, the average angular error is computed by randomly generating 20,000 normals per material and varying across all 100 material BRDFs in the database. This experiment is similar in setup to the one reported in [30] which, to our knowledge, is one of the most accurate techniques for photometric stereo on isotropic BRDFs. In [30], for 200 images, the angular error in estimating only the elevation angle *when the azimuth is known* is reported as $0.88°$; in contrast, our proposed technique, *without any prior knowledge of the azimuth*, has an angular error of $0.82°$ for the coarse-to-fine search and $0.58°$ for the gradient descent refinement.

**Varying BRDF.** Figure 5 compares the performance of the proposed technique to that of state-of-the art non-Lambertian photometric stereo algorithms [31], [32]. We fixed the number of images at $Q = 253$. Shown are aggregate statistics computed over 1,000 randomly-generated surface normals. For the coarse-to-fine technique, the worst-case error is less than $2°$ and, further, the error tapers down to $0.5°$ — which is the finest sampling of the candidate normals. Incorporating the gradient descent method provides substantial improvements and the angular error, in the best-case scenario, is reduced to $0.1°$, which is much smaller than the finest sampling on the candidate normals. This demonstrates the value of the gradient descent method over only coarse-to-fine search strategy. We also note that the proposed technique algorithm outperforms the both state-of-the-art techniques [31], [32] for most of materials we compare against; in total, the proposed technique has worse performance in 12 out of 100 materials. In addition to these simulations, in the supplemental material, we report the performance of the proposed normal estimation techniques as well as competing algorithms for multiple non-Lambertian BRDFs.

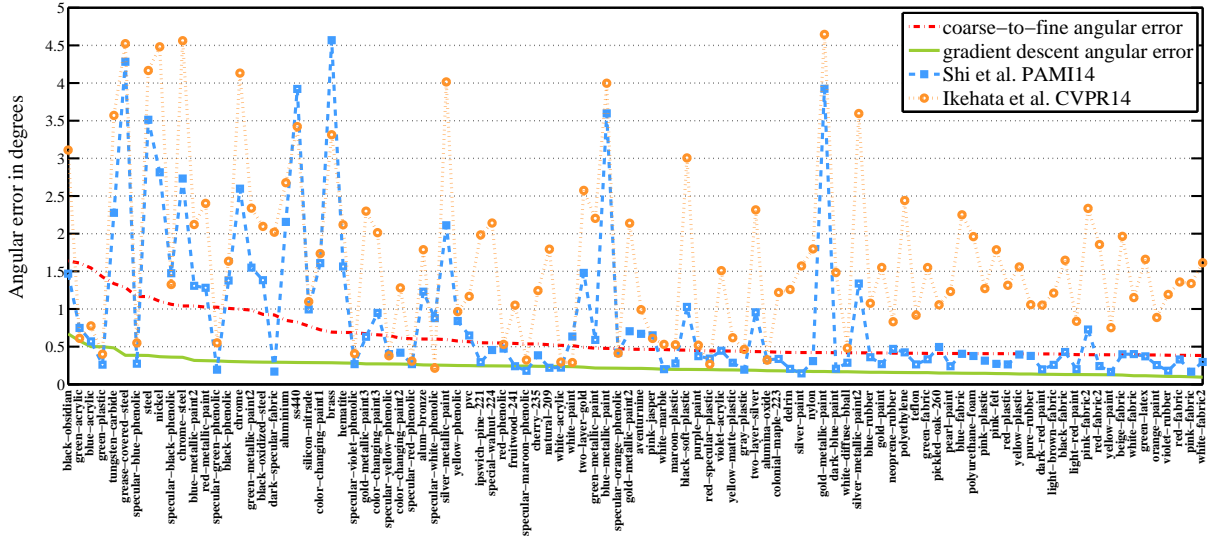

Fig. 5: **Normal estimation for different materials.** We fix the number of input images/lighting directions to 253. For each material BRDF, we compute average error over $1,000$ randomly-generated surface normals for both the coarse-to-fine search strategy (in red) and the gradient descent method (in green). The gradient descent scheme outperforms both competing methods [31], [32] in 88 out of 100 materials.

**Varying dictionary size and type.** Figure 6 evaluates the performance of both coarse-to-fine and gradient descent approaches as the number of dictionary atoms is varied. We use the same setting as Figure 5 but randomly pick 10, 30, 50, 70 and 90 atoms from the remaining 99 materials in MERL database. Shown are aggregate statistics computed over 5 trials. We observe that the angular errors of the gradient descent approach are less than $3°$ for most materials even for a small, randomly-selected dictionary.

Next, we evaluate the performance of our technique with specialized dictionaries that are comprised of BRDFs from similar materials. We construct three kinds dictionaries: (i) one each for paints, fabrics, plastics, phenolics, and metals; (ii) one dictionary whose atoms are randomly selected; and (iii) a leave-one-out dictionary made of all BRDFs except the one being tested. For evaluation, we isolate 10 materials — two each for the five categories in type (i) above — with no intersection between the training and test materials. Adopt the same setup from Figure 5 in terms of lighting directions and the number of input images, we evaluate the performance of the proposed technique on these 7 dictionaries in Figure 7 As is to be expected, a mismatch between the dictionary type and the test material produces unstable estimates. This trend is most distinct for metallic objects which have high-frequency components in their BRDFs. We also observe that the dictionaries with a mixture of materials returns the most stable performance. Finally, as expected, the leave-one-out dictionary with 99 atoms outperforms other dictionaries. This demonstrates the advantage of the proposed technique by using reference materials from a wide range.

#### 6.1.2 Performance of BRDF estimation

We characterize the performance of BRDF estimation by testing on the objects with spatial invariant as well as the spatially varying BRDFs.

Given a test BRDF, we generated 100 surface normals with random orientations and rendered their appearance for 253 lighting directions. Assuming the knowledge of the true surface normals, we estimate the BRDF using the optimization in Section 5, comparing the estimates produced by the per-pixel as well as low-rank constrained methods. We use the relative BRDF error [27] to quantify the accuracy of the estimate. Given true BRDF $\rho$ and estimated value $\widehat{\rho}$, the relative BRDF error is given as

$$\sqrt{\sum_i w_i ((\widehat{\rho}(i) - \rho(i)) \cdot \max(0, \cos(\theta_i)))^2 / \sum_i w_i}, \quad (8)$$

with $w_i$ set equal to 1 for convenience.

**Spatially-invariant BRDF.** Figure 4 characterizes the average relative BRDF error for varying number of lighting directions, which is computed by averaging all 100 material BRDFs in MERL database based on the $20,000$ random generated normals. Additional results on objects with spatially-invariant BRDF can be found in the supplemental material.

**Spatially-varying BRDF.** Though our model qualitatively and quantitatively performs well on the homogeneous ob-jects, objects with spatially-varying BRDF present a more challenging scenario. To illustrate this, we simulate an object whose SV-BRDF is constantly varying. An example is shown in Figure 8. We select three materials from the MERL database [26] and vary their relative abundances smoothly as shown in Figure 8. Now, the BRDF at each pixel in the rendered objects can be represented as a linear combination of the selected materials. In Figure 8, we showcase the performance of the low-rank BRDF estimation technique by rendering results obtained at different rank of the solution. We obtain solutions with varying ranks by tuning the value of $\beta$; for each value of the rank $K$, we show the corresponding value of $\beta$ at the top of the plot in Figure 8. The performance demonstrated in terms of both the qualitative results and quantitative measurements suggests robustness by incorporating low-rank prior. Note that the value of $K$ used for the optimization may not be consistent with the number of the underlying materials. That is, the BRDF at each pixel, which is the linear combination of selected BRDFs, may not be uniquely described by the BRDF dictionary due to linear correlation between the atoms. This naturally introduces a larger value of $K$ for the convergence of the relative BRDF errors.

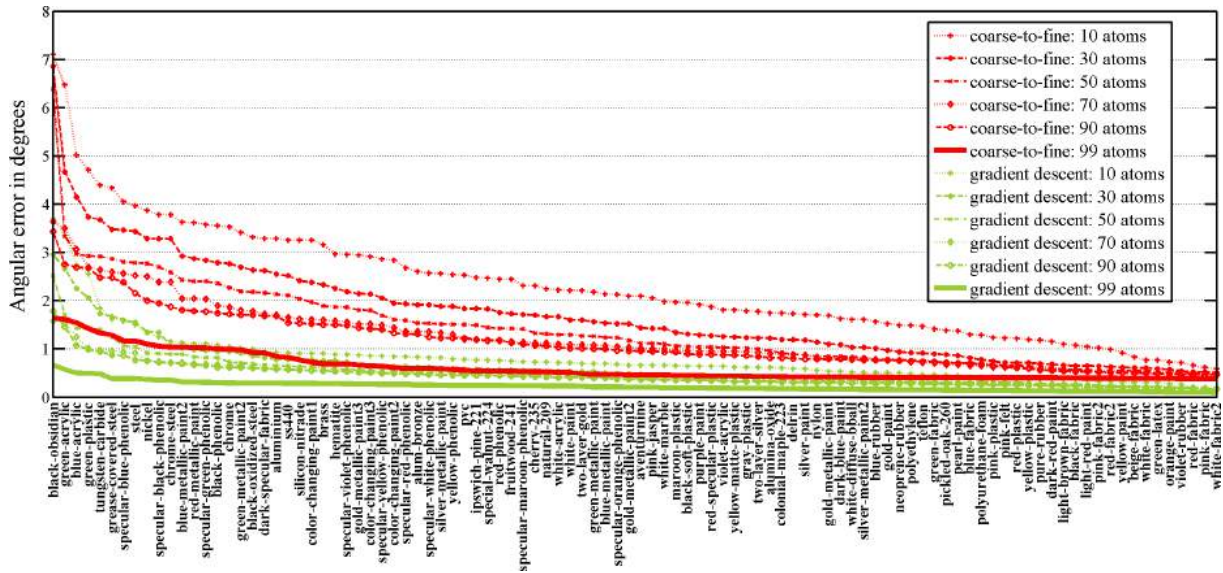

Fig. 6: **Normal estimation for different dictionary size.** We fix the number of input images/lighting directions to 253. For different dictionary size, we compute average error over 1,000 randomly-generated surface normals for both coarse-to-fine search (in red) and the gradient descent (in green) methods.

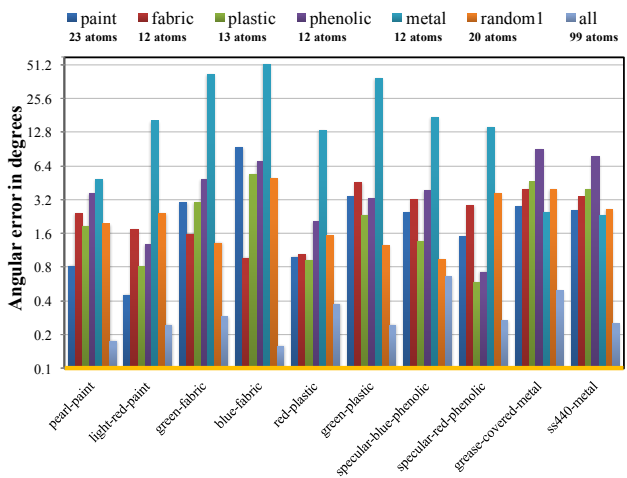

Fig. 7: **Normal estimation for different type of materials in the dictionary.** We fix the number of input images/lighting directions to 253. The numbers in the legend indicates the size of the corresponding dictionary. We observe that a mismatch in material type always leads to poor normal estimates. Shown are average errors over 1,000 randomly-generated surface normals.

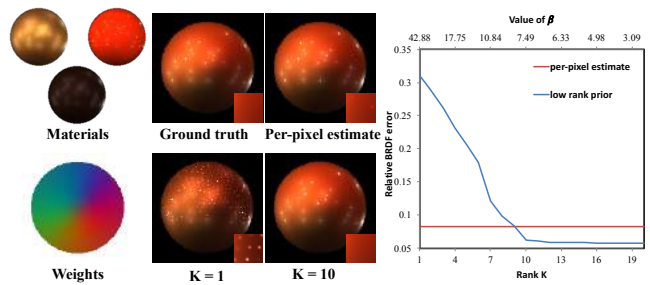

Fig. 8: **BRDF evaluation with low rank prior for a synthetic object.** We show a spherical object whose per-pixel BRDF is a linear combination of the three materials shown. The color coded sphere shows the relative abundances of the three materials in each color channel. We show rendered images using the ground truth, the per-pixel estimate as well as the low-rank estimate for different values of rank, $K$. For each value of the solution rank $K$, we include the corresponding value of $\beta$ used in the optimization at the top of the plot. Finally, we present the relative BRDF error as a function of the rank.

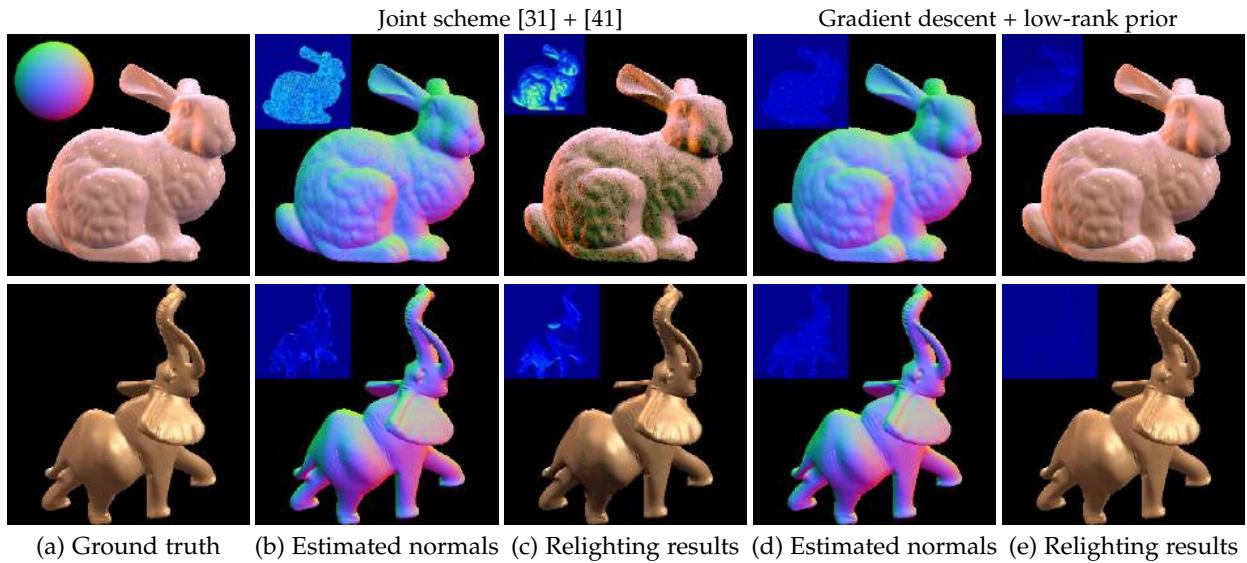

(a) Ground truth  (b) Estimated normals  (c) Relighting results  (d) Estimated normals  (e) Relighting results

Fig. 9: **Joint evaluation for the surface normals and BRDFs**. We compare on synthetic synthetic data for the proposed technique with the joint scheme of surface normals estimating using [31] and BRDF estimation using [41]. Insets on the top-left are the angular errors and euclidean intensity differences for the relighting, shown for both the joint scheme and the proposed approach. The proposed technique outperforms the competing methods in both normal estimation and novel image synthesis.

### 6.1.3 End-to-end performance characterization

To evaluate the end-to-end performance of both surface normal and BRDF estimation, we first characterize BRDF recovery using the estimated normals under varying number of light directions. Figure 4 characterizes the average relative BRDF errors for different number of light sources for the materials in the MERL database [26]. As seen here, the relative BRDF errors when using normal estimates of the gradient descent technique are in close proximity to errors when using the ground-truth surface normals. We also compare the relighting results for the proposed technique with the joint scheme with estimated surface normals from [31] and per-pixel BRDF fitting model from [41]. Specifically, given the surface normals estimated from [31], we perform BRDF fitting scheme as shown in [41]. We generate the test materials BRDFs by mixing materials from the MERL database and use 253 input images for the estimation. Given the estimated normals and BRDFs, we rendered the objects using the Grace Cathedral environment as the scene illumination. Figure 9 showcases the performance for the proposed technique and the combination of [31] and [41]. An evaluation of reconstruction error is shown in the top-left in the relighting results. We observe that the proposed technique outperforms the combination scheme for surface normal [31] and BRDF fitting [41] in terms of both visually results and quantitative measurements. While the methods [31] can produce good estimates for the surface normals and [41] can effectively address the BRDF fitting given the surface normals, they still need to solve the ill-posed problem by using priors to model the BRDFs or surface normals, making the framework fragile to the noisy estimates. In contrast, we solved for a sequence of well-defined problems, allowing for robust estimates for both surface normals and material BRDFs even noisy estimates present.

## 6.2 Real data

Real images are present a layer of difficulty well beyond simulations and introduce inter-reflections, sub-surface scattering, cast shadows, and imprecise light source localization. We test the performance of our shape and BRDF recovery algorithm on a wide range of datasets. Specifically, we use datasets from three sources — the benchmark dataset of [51], the light-stage data from [50], and the gourd from [7].

**Comparisons on benchmark dataset.** Figure 10 showcases the performance of non-Lambertian photometric stereo techniques on the benchmark dataset [51]. Each object in the database was captured with the ground truth surface normals, allowing for quantitative evaluations. For each object, we tabulate the mean and median of the angular errors for the estimated surface normals. The results for the benchmarked algorithms were done using code-base provided as part of the dataset.[3] For eight out of the ten objects, the gradient descent scheme outperforms all the methods provided as part of the benchmark in both mean and median angular error.

**Normal estimation.** Figure 11 showcases shape estimation of both the coarse-to-fine search and gradient descent method, respectively, on a variety of real datasets from the USC database [50]. We use Poisson reconstruction to obtain 3D surfaces from the estimated normal maps. The results were obtained from 250 input images. From the performance of the surface, the gradient descent method provides more fine scale structures as indicated in the red rectangle (the bolt on the shoulder-plate, the bolt on the helmet, the badge), as well as remove the artifacts shown in the `helmet`.

3. See https://sites.google.com/site/photometricstereodata/

| methods | metric | ball | cat | pot1 | bear | pot2 | buddha | goblet | reading | cow | harvest |
|---|---|---|---|---|---|---|---|---|---|---|---|
| WG10 [16] | mean | 2.03° | 6.72° | 7.18° | 6.50° | 13.12° | 10.91° | 15.70° | 15.39° | 25.89° | 30.00° |
| | median | 2.11° | 5.70° | 5.64° | 4.88° | 8.92° | 8.51° | 12.34° | 9.70° | 26.81° | 24.08° |
| IW12 [17] | mean | 2.54° | 7.21° | 7.74° | 7.32° | 14.09° | 11.11° | 16.25° | 16.16° | 25.70° | 16.11° |
| | median | 2.29° | 6.02° | 6.09° | 5.88° | 10.58° | 8.73° | 13.27° | 9.37° | 26.50° | 29.26° |
| GC10 [8] | mean | 3.21° | 8.22° | 8.53° | 6.62° | 7.90° | 14.85° | 14.22° | 19.07° | 9.55° | 27.84 |
| | median | 1.17° | 4.67° | 4.01° | **3.61°** | 3.37° | 7.57° | 8.01° | 14.07° | 5.79° | 20.22° |
| AZ08 [9] | mean | 2.71° | 6.53° | 7.23° | 5.96° | 11.03° | 12.54° | 13.93° | 14.17° | 21.48° | 30.51° |
| | median | 2.47° | 4.32° | 4.70° | 3.97° | 8.40° | 7.62° | 9.64° | 7.23° | 21.52° | 18.34° |
| HM10 [37] | mean | 3.55° | 8.40° | 10.85° | 11.48° | 16.37° | 13.05° | 14.89° | 16.82° | 14.95° | 21.79° |
| | median | 2.86° | 6.07° | 7.35° | 9.81° | 13.07° | 9.14° | 10.10° | 11.34° | 12.70° | 14.88° |
| ST12 [34] | mean | 13.58° | 12.33° | 10.37° | 19.44° | 9.84° | 18.37° | 17.80° | 17.17° | **7.62°** | 19.30° |
| | median | 12.32° | 9.57° | 7.52° | 19.07° | 6.67° | 15.48° | 14.04° | 12.74° | **3.91°** | 13.58° |
| ST14 [35] | mean | 1.74° | 6.12° | 6.51° | 6.12° | 8.78° | 10.60° | 10.09° | 13.63° | 13.93° | 25.44° |
| | median | 1.57° | 4.04° | 4.05° | 4.38° | 6.50° | 6.89° | 7.27° | 7.59° | 12.17° | 17.12° |
| IA14 [36] | mean | 3.34° | 6.74° | 6.64° | 7.11° | 8.77° | 10.47° | 9.71° | 14.19° | 13.05° | 25.95° |
| | median | 3.33° | 4.86° | 4.24° | 5.57° | 6.57° | 6.71° | 6.59° | 8.21° | 10.59° | 17.40° |
| Gradient descent | mean | **1.33°** | **4.88°** | **5.16°** | **5.58°** | **6.41°** | **8.48°** | **7.57°** | **12.08°** | 8.23° | **15.81°** |
| | median | **0.91°** | **3.04°** | **2.55°** | 4.45° | **3.18°** | **5.36°** | **5.10°** | **5.35°** | 4.58° | **7.74°** |

Fig. 10: **Evaluations on the benchmark photometric stereo dataset [51].** Shown are the mean and median of the angular errors measured in degrees for both the gradient descent method and the state-of-the-art techniques. For each object, the best performing algorithm for both mean and median angular error is marked in red. The proposed technique outperforms the benchmarked techniques in a majority of scenes. The numbers for the benchmarked algorithms are reported from [51].

**BRDF estimation.** Next, we showcase the performance of BRDF estimation on the `knight` scene using the surface normals estimated using the gradient descent technique. The object in the scene exhibits many unique materials (the helmet, the breast-plate, the chain, the red scabbard, to name a few) as well as significant modeling deviations (inter-reflections, cast-shadows). Figure 12 shows rendered photographs under natural lighting based on the USC light probes [52] for both the per-pixel and low-rank prior approaches. While the per-pixel estimates show the robustness to handle objects with complex spatial variations, it produces noisy rendering results due to insufficient observations. Incorporating the low-rank prior returns a more faithful rendition of the scene, indicating the advantages gained by pooling the information across multiple pixels.

**Evaluations.** Figure 13 showcases the performance of our algorithm on two real datasets (`gourd1` and `helmet`). The results for the `helmet` were obtained from 250 input images, and the results of `gourd1` were obtained from 100 input images. The recovered shape and BRDF (as visualized via rendered images) seem to be in agreement with the results in [7]; however, our algorithm is significantly simpler and employs a per-pixel algorithm that be easily parallelized. The proposed estimation framework showcases its robustness to handle objects with complex spatial varying materials and render faithful renditions under both simple and complex lighting environment. We refer the reader to the supplemental videos highlighting the relighting results.

## 7 DISCUSSIONS

We present a photometric stereo technique for per-pixel normal and BRDF estimation for objects that are visually complex. We demonstrate that the use of a BRDF dictionary significantly simplifies the inverse problem and provides not just state-of-the-art results in normal and BRDF estimation but also works robustly on a wide range of real scenes. The hallmark of our approach is the ability to obtain surface normal and SV-BRDF estimates without requiring complex iterative techniques endemic to state-of-the-art techniques [6], [7]. Finally, our per-pixel estimation framework is ripe for further speed-ups by solve for the shape and reflectance at each pixel in parallel.

**Limitations.** While the use of virtual exemplars provides flexibility beyond [10], we require light calibration and hence, our method is most suited to shape and reflectance acquisition from light-stages where the light sources are fixed and the calibration is a one-time effort. Finally, it is also important that the scene lies in the linear span of our dictionary. In the failure of this, our results can be unpredictable. Here, the need for a larger dictionary encompassing hundreds, if not thousands, of materials would be invaluable for the broader applicability of our method.

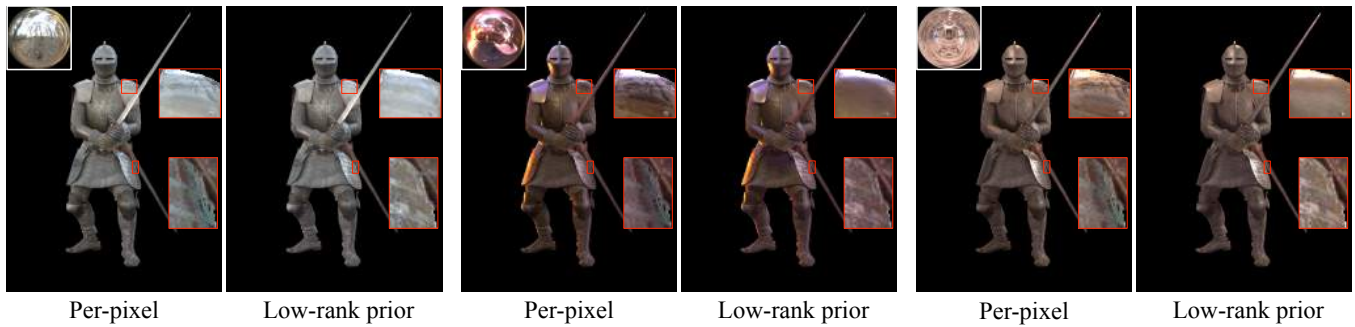

Per-pixel  Low-rank prior  Per-pixel  Low-rank prior  Per-pixel  Low-rank prior

Fig. 12: **Natural environment relighting results on `knight` dataset.** (From left to right) Shown are the rendering results under Eucalyptus Grove, Grace Cathedral and St. Peter's Basilica environment [52] for both the per-pixel and low-rank prior approaches. We also show the close-up appearance for the relighting results to highlight the improvements by incorporating the low-rank prior. Note how the shadings on the shoulder blades and thigh plates are correctly rendered by the low-rank prior.

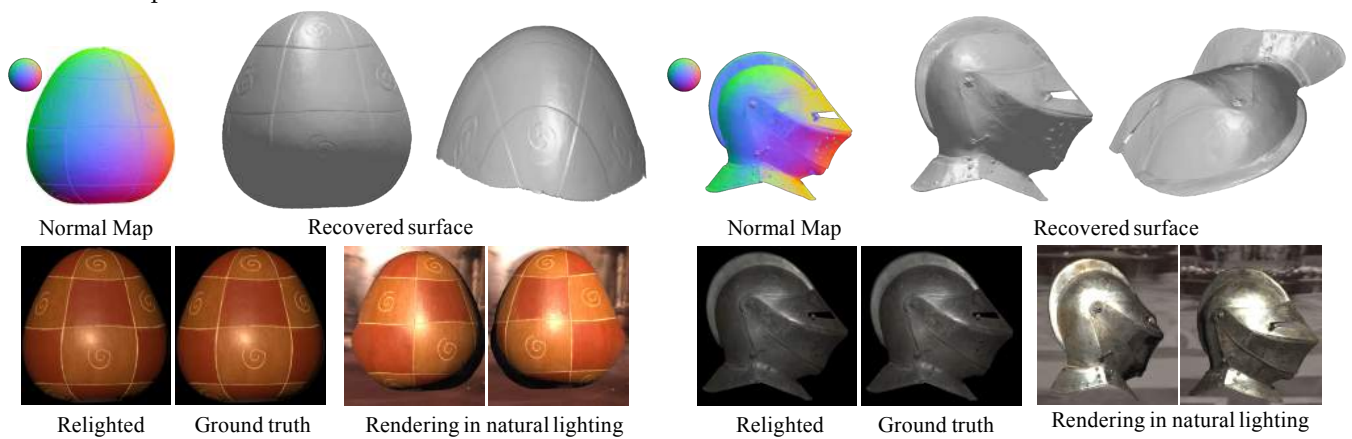

Normal Map  Recovered surface  Normal Map  Recovered surface

Relighted  Ground truth  Rendering in natural lighting  Relighted  Ground truth  Rendering in natural lighting

Fig. 13: **Results on `gourd1` and `helmet` datasets.** For both datasets, we show the estimated normal map in false color (top-left) and 3D surface (top-right) recovered from it. We also show the relighting results (bottom-left), ground truth under the same lighting direction (bottom-middle), and relighting under natural environment (bottom-right).

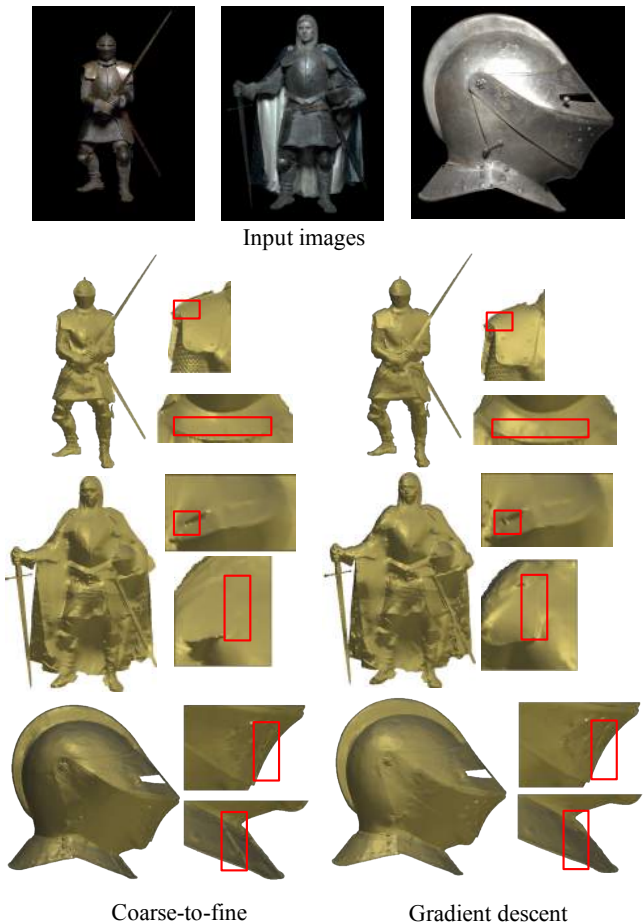

Fig. 11: **Recovered surfaces on several real scenes with complex, spatially varying reflectance.** Shown are the sample images for scene from the USC light-stage database [50] and 3D surfaces obtained by using Poisson reconstruction on the estimated normal maps. We also highlight differences between the coarse-to-fine and gradient descent approaches using red boxes. We observe that the gradient descent technique is able to preserve subtle details.

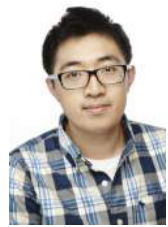

**Zhuo Hui** received his B.Eng. degree in electronics and information engineering in 2011 from The Hong Kong Polytechnic University. He is currently a Ph.D. student in electrical and computer engineering at Carnegie Mellon University (CMU). His research interests include computational photography and computer graphics.

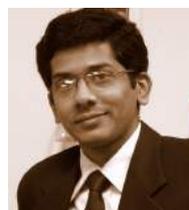

**Aswin C. Sankaranarayanan** is an Assistant Professor at the ECE Department in the Carnegie Mellon University (CMU) and the PI of the Image Science Lab. He received his Ph.D. from University of Maryland, College Park where he was awarded the distinguished dissertation fellowship for his thesis work by the ECE department in 2009. He was a post-doctoral researcher at the DSP group at Rice University. Aswin's research encompasses problems in compressive sensing and computational imaging. He has received best paper awards at the CVPR Workshops on Computational Cameras and Displays (2015) and Analysis and Modeling of Faces and Gestures (2010) as well as the Herschel Rich Invention Award from Rice University in 2016.